\documentclass{article}

\usepackage[preprint, nonatbib]{CRL_SR}

\usepackage[utf8]{inputenc} % allow utf-8 input
\usepackage[T1]{fontenc}    % use 8-bit T1 fonts
\usepackage{hyperref}     % hyperlinks
\usepackage{url}            % simple URL typesetting
\usepackage{booktabs}       % professional-quality 

\usepackage{amsfonts}       % blackboard math symbols
\usepackage{nicefrac}       % compact symbols for 1/2, etc.
\usepackage{microtype}      % microtypography
\usepackage{xcolor}         % colors
\usepackage{cite}
\usepackage{graphicx}
\usepackage{epsfig,epstopdf}
\usepackage{multirow}
\newcommand{\tabincell}[2]{\begin{tabular}{@{}#1@{}}#2\end{tabular}}
\usepackage{bm}

\title{Blind Image Super-Resolution via Contrastive Representation Learning}

\author{%
  Jiahui Zhang \qquad Shijian Lu \thanks{corresponding author} \qquad Fangneng Zhan \qquad Yingchen Yu\\
  School of Computer Science and Engineering\\
  Nanyang Technological University\\
  \texttt{JIAHUI003@e.ntu.edu.sg}, \qquad \texttt{Shijian.Lu@ntu.edu.sg},  \\ 
  \texttt{fnzhan@ntu.edu.sg}, \qquad \texttt{yingchen001@e.ntu.edu.sg}  \\
}

\begin{document}

\maketitle

\begin{abstract}
Image super-resolution (SR) research has witnessed impressive progress thanks to the advance of convolutional neural networks (CNNs) in recent years. However, most existing SR methods are \textit{non-blind} and assume that degradation has a single fixed and known distribution (e.g., bicubic) which struggle while handling degradation in real-world data that usually follows a multi-modal, spatially variant, and unknown distribution. The recent \textit{blind SR} studies address this issue via 
degradation estimation, but they do not generalize well to multi-source degradation and cannot handle spatially variant degradation. We design CRL-SR, a contrastive representation learning network that focuses on blind SR of images with multi-modal and spatially variant distributions. CRL-SR addresses the blind SR challenges from two perspectives. The first is contrastive decoupling encoding which introduces contrastive learning to extract resolution-invariant embedding and discard resolution-variant embedding under the guidance of a bidirectional contrastive loss. The second is contrastive feature refinement which generates lost or corrupted high-frequency details under the guidance of a conditional contrastive loss. % CRL-SR achieve accurate and robust SR in a two-stage manner. The first stage tackles decoupling encoding which introduces contrastive learning to extract resolution-invariant embedding and discard resolution-variant embedding under the guidance of a bidirectional contrastive loss. The second stage focuses on feature refinement which generates lost or damaged high-frequency details under the guidance of a conditional contrastive loss. 
Extensive experiments on synthetic datasets and real images show that the proposed CRL-SR can handle multi-modal and spatially variant degradation effectively under blind settings and it also outperforms state-of-the-art SR methods qualitatively and quantitatively.
\end{abstract}

\section{Introduction}
\label{introduction}
Image super-resolution (SR) aims to reconstruct an accurate high-resolution (HR) image from its low-resolution (LR) counterpart. It is a typical ill-posed problem as the LR-to-HR mapping is naturally nondeterministic. Thanks to the development of convolutional neural networks (CNNs) in recent years, quite a number of SR methods \cite{li2019gated, kim2016deeply, he2019ode, dai2019second, li2018multi, li2019feedback, haris2018deep, lai2017deep, zhang2018image, zhang2018residual, lim2017enhanced, guo2020closed, xu2020unified, mei2020image, liu2020residual} exploits CNNs and have achieved very impressive SR performance. However, most existing methods assume that the degradation in LR images follows some single fixed and known distribution (e.g., bicubic) which struggle while handling real-world data that often have multi-modal, spatially variant and unknown degradation distribution. High-fidelity SR under the presence of degradation with multi-modal, spatially variant and unknown distribution is still an open research challenge.

Quite a number of non-blind SR methods \cite{zhang2018learning, xu2020unified, shocher2018zero, soh2020meta, zhang2020deep} have been proposed for handling multi-modal image degradation as usually borne in real-world data. For example, Zhang et al. \cite{zhang2018learning} exploit the types of degradation (including blur kernels and noises) as certain priors and train a single network for modelling multi-modal degradation. Recently, Xu et al. \cite{xu2020unified} introduce dynamic convolutions into the SR task for modelling spatially variant degradation within the same or across different images. Nonetheless, these non-blind methods share a common assumption that degradation in LR images follows some known distribution which is usually inapplicable in real-world data.

Several blind methods have been reported for modelling degradation with unknown distributions. Inspired by the phenomenon of kernel mismatch, Gu et al. \cite{gu2019blind} propose an iterative kernel correction technique for blur kernel estimation for blind SR. Although the iterative kernel correction performs better than direct kernel estimation, it tends to suffer from fair estimation errors while dealing with complex multi-modal spatially-variant degradation. In addition, Kligler et al. \cite{bell2019blind} propose blind KernelGAN that exploits self-similarity of natural images to estimate SR kernels. It is susceptible to prediction errors as well while handling multi-modal spatially-variant degradation.

We propose CRL-SR, a contrastive representation learning network that aims to address the blind SR challenge while the image degradation follows certain multi-modal, spatially variant, and unknown distributions. Leveraging the idea of the recent instance contrastive learning \cite{chen2020simple, tian2019contrastive, he2020momentum}, we design a novel contrastive decoupling encoding (CDE) technique that employs a pair of encoders to encode \textit{resolution-invariant} features (across LR and HR images) only while discarding \textit{resolution-variant} features from LR images. Since a fair amount of high-frequency features are lost in LR images, we define all low-frequency features and those well-kept high-frequency features in LR images as \textit{resolution-invariant}, and degradation (e.g. blurs and noises) in LR images and the lost high-frequency features in HR images as \textit{resolution-variant}. CDE thus strives to maximize the mutual information between LR and HR images which greatly facilitates the reconstruction of the resolution-invariant features under the guidance of a novel bidirectional contrastive loss. In addition, we propose a contrastive feature refinement (CFR) technique that recovers the lost high-frequency details for high-fidelity SR. With the reconstructed resolution-invariant features as priors, we design a conditional contrastive loss that guides to generate high-frequency details and pulls the reconstructed SR images towards HR images in the feature representation space. Thanks to the contrastive decoupling and generation designs, the proposed CRL-SR can handle multi-modal and spatially variant degradation and reconstruct superior image structures and fine details.

The contributions of this work can be summarized in three aspects. First, we design CRL-SR, a contrastive representative learning network that offers a new and effective blind SR approach while images suffer from multi-modal, spatially variant, and unknown degradation. %Extensive experiments show that CRL-SR achieves superior SR performance qualitatively and quantitatively. 
Second, we design a contrastive decoupling encoding technique that leverages feature contrast to extract resolution-invariant features across LR and HR images which effectively mitigates the adverse effects of resolution-variant degradation in SR reconstruction. Third, we design a conditional contrastive loss that guides to generate the high-frequency details that are lost during the image degradation process effectively.

\section{Related Work}

\textbf{CNN-based Methods for Single Degradation.} 
Convolutional Neural Network (CNN) have been investigated extensively in various image generation tasks such as image translation \cite{park2019spade,isola2017pix2pix,zhan2019gadan,zhan2020sagan,zhan2021unite,zhan2021rabit,zhan2019esir},
image editing \cite{yu2018inpainting,wu2020cascade,zhan2020aicnet,wu2020leed,zhan2020towards}, 
image composition \cite{lin2018st,zhan2019sfgan,zhan2021emlight,zhan2021gmlight,zhan2021needlelight,cui2021fbcgan,zhan2018verisimilar,zhan2019scene}, 
image inpainting \cite{iizuka2017globally,yu2019free,yu2021diverse} etc.
Specially, CNN-based image SR methods have achieved very impressive performance while image degradation follows a single and known distribution. Leveraging the pioneer work \cite{dong2014learning} that adopts CNNs for SR, \cite{kim2016accurate} and DRCN \cite{kim2016deeply} introduce residual learning that allows a larger number of network layers and improves SR greatly. Some work focuses on extracting better deep network features, e.g. \cite{lim2017enhanced} and \cite{lim2017enhanced} introduce residual scaling for this purpose. Besides, several methods exploit multi-scale hierarchical features from each instead of just the last CNN layer. For example, \cite{zhang2018residual} presents a residual dense network to make full use of features from all Conv layers, \cite{zhang2018image} presents a very deep residual channel attention network that introduces channel attention, and \cite{dai2019second} considers feature correlation via second-order channel attention. However, the aforementioned methods all assume a single and fixed degradation model which cannot handle real-world images that often suffer from much more complicated degradation.

\textbf{CNN-based Methods for Multiple Degradation. } SR under the presence of multiple degradation models has attracted increasing attention in recent years. In particular, \cite{zhang2018learning} employs a single network to handle different types of degradation which uses blur kernels and noise levels of the degradation as priors in network training. \cite{xu2020unified} introduces dynamic convolutions with per-pixel kernels %\cite{xu2020unified}
to handle multiple degradation models with spatially variant distributions. However, these methods are non-blind which require certain prior knowledge of the degradation in both network training and inference. Several blind SR methods have been reported recently since degradation models in real-world data are usually unknown. For example, \cite{gu2019blind} implements multiple iterations to predict degradation kernels and combines the predicted kernels with spatial feature transform \cite{gu2019blind} for blind SR. \cite{bell2019blind} also predicts kernels and it exploits the internal cross-scale recurrence \cite{bell2019blind, shocher2018zero} of images to estimate degradation kernels. Most of these blind SR methods assume a constant degradation kernel throughout the whole image which often introduces kernel prediction errors and further SR artifacts and blurs while the image degradation is spatially variant at different image locations.

We focus on image degradation that has multiple spatially variant distributions with unknown distribution models and parameters. To the best of our knowledge, this type of degradation is closest to the real-world degradation that has been studied in the prior blind SR research.

\textbf{Contrastive Learning. } Contrastive learning has been widely studied for unsupervised representation learning in recent years. It works by pulling positive samples close to the anchor and pushing negative samples away in the representation space, hence increases the mutual information in the learned representation \cite{sermanet2018time,tian2019contrastive,chen2020simple,he2020momentum,oord2018representation}. The selection of the positive and negative samples depends on specific downstream tasks. For example, \cite{chen2020simple, he2020momentum} treat the augmentation of the original data as positive samples. \cite{tian2019contrastive, sermanet2018time} consider multiple views of the same sample as positive pairs. 

We leverage contrastive learning in two tasks in the proposed CRL-SR. First, we adopt contrastive learning in the decoupling encoder for extracting common (resolution-invariant) features from HR and LR which mutually form positive pairs. Second, we adopt contrastive learning in the feature refinement for recovering the lost high-frequency details by pulling SR reconstruction (positive) towards HR images (anchor). More details about the selection of positive and negative samples and the design of the two contrastive losses are to be discussed in Section ~\ref{Proposed Method}.

\begin{figure*}[tbp]
\begin{center}
\includegraphics[width=1\textwidth]{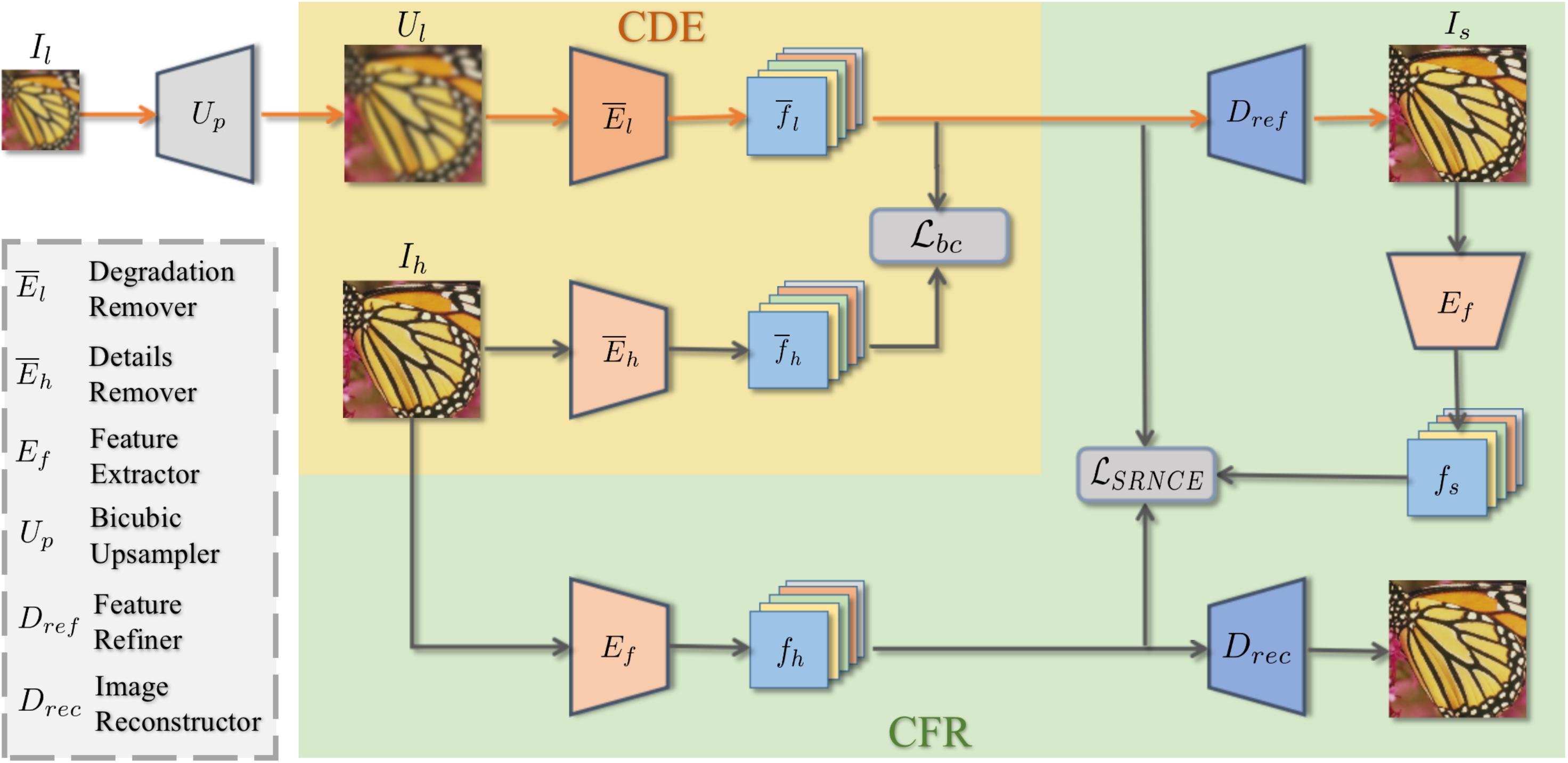}
\end{center}
\caption{
The proposed CRL-SR addresses the blind SR challenges through contrastive decoupling encoding and contrastive feature refinement. CDE involves a pair of encoders {$\overline{E}_{l}$} and {$\overline{E}_{h}$} that jointly learn resolution-invariant features {$\overline {f}_{h}$} and {$\overline{f}_{l}$} by removing resolution-variant degradation and high-frequency details from LR image $I_{l}$ and HR image $I_{h}$ under the guidance of a bidirectional contrastive loss {$\mathcal{L}_{bc}$}. CFR tackles feature refinement with {$D_{ref}$} that generates the lost high-frequency details under the guidance of a conditional contrastive loss {$\mathcal{L}_{SRNCE}$}. {$E_{f}$} extracts features {$f_{h}$} and {$f_{s}$} from {$I_{h}$} and the reconstructed image {$I_{s}$}, and {$D_{rec}$} is a decoder that reconstructs the high-resolution image. The processes linked by orange arrows are involved in both training and inference, and those linked by grey arrows are involved in training only.}
\label{model}
\end{figure*}

\section{Proposed Method}
\label{Proposed Method}
\subsection{Problem Formulation}

With the target of tackling \textit{real-world degradation} which usually comes with multi-modal, spatially variant, and unknown distributions, 
we apply blurs (with globally constant or spatially variant distributions) to HR images and then down-sample them and include noises to obtain LR images. Following \cite{wang2021unsupervised, gu2019blind, xu2020unified}, the whole degradation process can be formulated by:
\begin{equation}
\label{eq1}
I_{l}=(I_{h}\otimes{k})\downarrow_{s}+n,
\end{equation}
where {I$_{h}$} denotes the HR image, {$k$} denotes a blur kernel, {$\otimes$} denotes convolution operations, {$\downarrow_{s}$} stands for down-sampling operation with a scale factor {$s$}, and {$n$} represents additive white Gaussian noises. We adopt bicubic down-sampler to down-sample images as in \cite{zhang2018residual, xu2020unified, li2019feedback, zhang2018learning, gu2019blind}. We assume little knowledge known about the degradation parameters (in both CRL-SR designs and the ensuing experiments) for approximating real-world data and degradation as much as possible.

\subsection{Overview}

The proposed CRL-SR tackles blind SR from a new perspective of contrastive representation learning. It achieve blind SR with two novel designs including contrastive decoupling encoding and contrastive feature refinement. As illustrated in Fig.~\ref{model}, CDE consists of two encoders {$\overline{E}_{l}$} and {$\overline{E}_{h}$} which learns to remove degradation information from the up-sampled LR images {$U_{l}$} and remove high-frequency details in HR images {$I_{h}$}, respectively. The two encoders are trained under the guidance of a bidirectional contrastive loss {$\mathcal{L}_{bc}$} which strives to disentangle and keep resolution-invariant features (i.e., clean low-frequency features) and discard resolution-variant features. More details about the CDE are to be presented in the ensuing Section \ref{Contrastive Decoupling Encoding Scheme}.

With the disentangled resolution-invariant features ({$\overline{f}_{l}$}), CFR is designed to recover high-frequency details that are lost or corrupted during the image degradation process. 
Specifically, a feature generation module {$D_{ref}$} is employed to generate the high-frequency details and reconstruct the SR image {$I_{s}$}. To ensure the reconstruction quality, we propose a novel conditional contrastive loss {$\mathcal{L}_{SRNCE}$} which takes {$\overline{f}_{l}$}, {$f_{h}$} and {$f_{s}$} (i.e. the features of {$U_{l}$}, {$I_{h}$} and {$I_{s}$}) as inputs and guides to maximize the mutual information between the generated {$I_{s}$} and {$I_{h}$} in the feature representation space. More details about the CFR are to be presented in the ensuing Section~\ref{Contrastive Feature Refinement Scheme}.

\subsection{Contrastive Decoupling Encoding}
\label{Contrastive Decoupling Encoding Scheme}
We design the CDE to disentangle resolution-invariant features (i.e. low-frequency features) across HR and LR domains. With the up-sampled image {$U_{l}$} and HR image {$I_{h}$} as inputs, we design two encoders {$\overline{E}_{l}$} and {$\overline{E}_{h}$} that strive to maximize the mutual information between their extracted features {$\overline{f}_{h}$} and {$\overline{f}_{l}$} as driven by a bidirectional contrastive loss {$\mathcal{L}_{bc}$}. 
The two encoders thus jointly extract resolution-invariant features and discard resolution-variant features (i.e., degradation of LR images and high-frequency details of HR images) simultaneously. 
Note that both {$\overline{f}_{h}$} and {$\overline{f}_{l}$} consist of \textit{M} feature vectors that are extracted by { $\overline{E}_{h}$} and {$\overline{E}_{l}$},  respectively. 

The bidirectional contrastive loss {$\mathcal{L}_{bc}$} has two components {$\mathcal{L}_c(\overline{f}_{h}, \overline{f}_{l})$} and {$\mathcal{L}_c(\overline{f}_{l}, \overline{f}_{h})$}, where the first argument of $\mathcal{L}_c(\cdot)$ is anchor and the second argument contains positive and negative samples. 
Take {$\mathcal{L}_c(\overline{f}_{h}, \overline{f}_{l})$} as an example. It treats each of the \textit{M} feature vectors in {$\overline{f}_{h}$} as an anchor ({$\overline{f}^{m}_{h}, m\!\in\!\mathbb{R}^{M})$}, the corresponding feature vector in {$\overline{f}_{l}$} as the positive sample ($\overline{f}^{m}_{l}, m\!\in\!\mathbb{R}^{M}$), and the remaining (\textit{M}-1) feature vectors in $\overline{f}^{m}_{l}$ as negative samples. We use the remaining (\textit{M}-1) feature vectors as negative samples as they are more difficult to distinguish from the anchor (than features from other images) due to internal image statistics \cite{shocher2018zero, bell2019blind, glasner2009super}. {$\mathcal{L}_c(\overline{f}_{h}, \overline{f}_{l})$} can thus be formulated based on the noise contrastive estimation framework \cite{oord2018representation}:
\begin{equation}
\mathcal{L}_c(\overline{f}_{h}, \overline{f}_{l})= -\sum_{m=1}^M\log \Bigg [\frac{e^{{\overline{f}^{m}_{h}} \cdot \overline{f}^{m}_{l} \frac{1}{\tau}}}{\sum_{j=1}^M e^{{\overline{f}^{m}_{h}}\cdot \overline{f}^{j}_{l} \frac{1}{\tau}}}\Bigg ]
\end{equation}

where $\tau$ denotes a temperature hyper-parameter and `$\cdot$' stands for the dot product of two feature vectors. Note before the loss computation, all feature vectors are first mapped to \textit{S}-dimensional feature vectors in a higher dimensional space by a two-layer perceptron (MLP) as in \cite{chen2020simple}. After that, the mapped feature vectors are normalized to stabilize the training process. The overall bidirectional contrastive loss {$\mathcal{L}_{bc}$} can thus be formulated by:
\begin{equation}
\mathcal{L}_{bc} = \mathcal{L}_c(\overline{f}_{h}, \overline{f}_{l}) + \mathcal{L}_c(\overline{f}_{l}, \overline{f}_{h})
\end{equation}

\subsection{Contrastive Feature Refinement}
\label{Contrastive Feature Refinement Scheme}

Direct generation of the lost high-frequency details is challenging as the learned resolution-invariant features contain little relevant information. We design contrastive feature refinement to restore the lost high-frequency details. CFR consists of three components as illustrated in Fig. \ref{model}, which includes $D_{ref}$ that generates the lost high-frequency details, two $E_{f}$ that map HR and SR images to a feature space, and $D_{rec}$ that reconstructs HR images from this feature space. 

\begin{figure*}[!htbp]
\begin{center}
\includegraphics[width=1\textwidth]{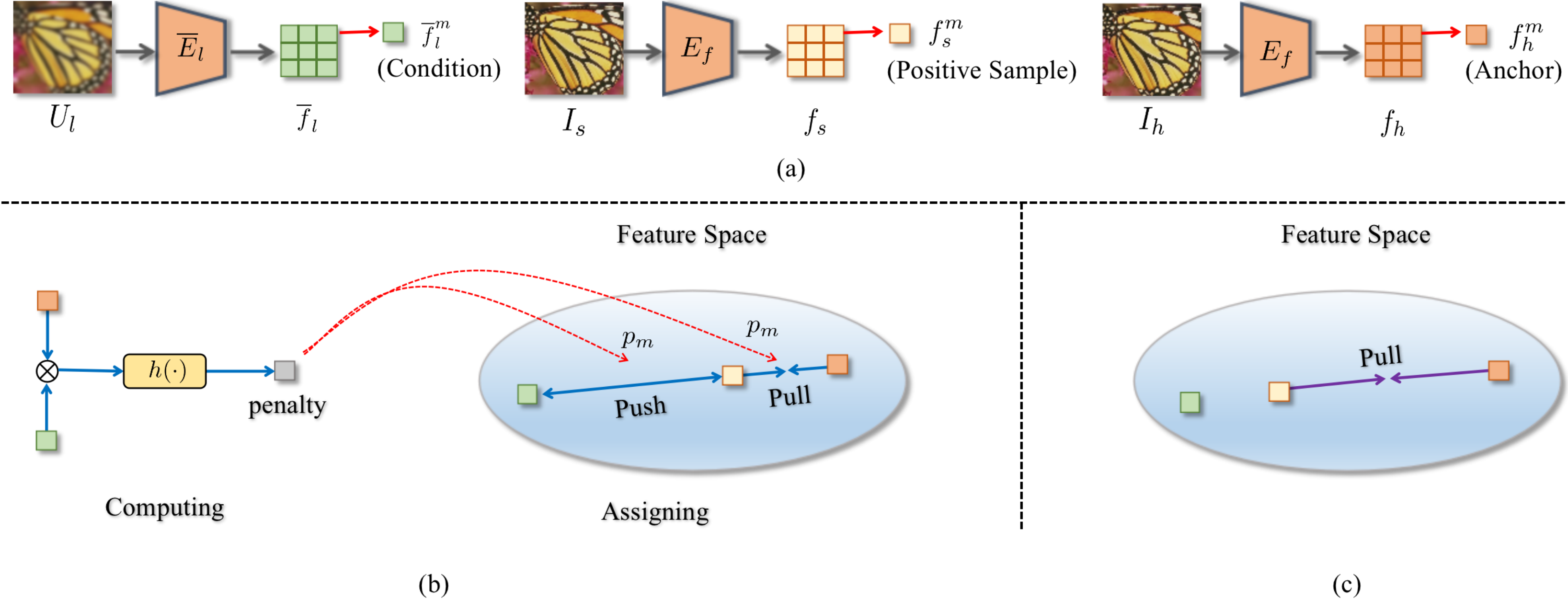}
\end{center}
\caption{
%The proposed conditional contrastive loss: Given the generated \textit{condition} (i.e. resolution-invariant LR features), positive sample, and anchor in (a), the proposed conditional contrastive loss takes $\overline{f}^{m}_{l}$ as condition to generate an extra pushing force to move $f^{m}_{s}$ towards {$f^{m}_{h}$}. In addition, this condition helps compute a penalty {$p_{m}$} for identifying ($\overline{f}^{m}_{l}$, $f^{m}_{h}$) pairs that are far apart and guiding the model pay more attention to the discrepancy of the corresponding pair ($\overline{f}^{m}_{l}$, $f^{m}_{s}$) and the compactness of the pair ({$f^{m}_{h}$}, {$f^{m}_{s}$}) as in (b). InfoNCE does not consider {$\overline{f}^{m}_{l}$} and it only pulls {$f^{m}_{s}$} towards {$f^{m}_{h}$} as in (c).
Illustration of the proposed conditional contrastive loss: With the resolution-invariant LR feature $\overline{f}^{m}_{l}$ in (a) as \textit{condition}, the proposed conditional contrastive loss introduces an extra pushing force between $\overline{f}^{m}_{l}$ and $f^{m}_{s}$ for pushing the \textit{positive sample} $f^{m}_{s}$ towards the \textit{anchor} {$f^{m}_{h}$}. Since different $\overline{f}^{m}_{l}$ loses different amount of high-frequency details, we exploit this condition and derive a penalty $p_{m}$ based on the distance between $\overline{f}^{m}_{l}$ (a measure of how much high-frequency detail is lost in each LR feature vector) and $f^{m}_{h}$ which adjusts the pushing force between $\overline{f}^{m}_{l}$ and $f^{m}_{s}$ as well as the pulling force between $f^{m}_{s}$ and $f^{m}_{h}$ adaptively as illustrated in (b). InfoNCE does not consider {$\overline{f}^{m}_{l}$} and it only pulls {$f^{m}_{s}$} towards {$f^{m}_{h}$} as shown in (c).
}
\label{ccloss}
\end{figure*}

We design a novel conditional contrastive loss that conditions on the resolution-invariant features $\overline{f}_{l}$ for effective detail generation and feature refinement. In the conditional contrastive loss, each of the \textit{M} HR feature vectors $f^{m}_{h}(m\!\in\!\mathbb{R}^{M})$, i.e. the \textit{M} spatial locations of each layer, serves as an anchor, the corresponding SR feature vector $f^{m}_{s}(m\!\in\!\mathbb{R}^{M})$ serves as the positive sample, and the remaining (\textit{M}-1) SR feature vectors $f^{j}_{s} (j\!\in\!\mathbb{R}^{M-1})$ serve as negative samples. Fig.~\ref{ccloss}a illustrates how the three types of features are extracted from $U_l$, $I_s$, and $I_h$.

%We adopt the resolution-invariant LR features as conditions since we aim to push the reconstructed SR features towards the corresponding HR features as close as possible. As Fig.~\ref{ccloss} shows, InfoNCE only considers the pulling force between {$f^{m}_{s}$} and {$f^{m}_{h}$}, whereas the proposed conditional contrastive loss introduces an extra pushing force between {$\overline{f}^{m}_{l}$} and {$f^{m}_{s}$} conditioning on {$\overline{f}^{m}_{l}$}. Specifically, we include the discrepancy of ( {$\overline{f}^{m}_{l}$}, {$f^{m}_{s}$}) (i.e. {$1 - {f^{m}_{s}}\cdot \overline{f}^{m}_{l}$}) to the InfoNCE. It can be formulated by:
The resolution-invariant LR feature {$\overline{f}^{m}_{l}$} affects the conditional contrastive loss in two different manners. First, it introduces an extra pushing force that pushes the reconstructed SR features towards the corresponding HR features as close as possible. Specifically, we include a pushing force between {$\overline{f}^{m}_{l}$} and {$f^{m}_{s}$} which strives to push {$f^{m}_{s}$} away from {$\overline{f}^{m}_{l}$} as illustrated Fig.~\ref{ccloss}b (within the elliptic feature space). This is unlike the standard InfoNCE that only considers the pulling force between {$f^{m}_{s}$} and {$f^{m}_{h}$} as illustrated in Fig.~\ref{ccloss}c. By including the pushing force between {$\overline{f}^{m}_{l}$} and {$f^{m}_{s}$} (i.e. {$1 - {f^{m}_{s}}\cdot \overline{f}^{m}_{l}$}), a new contrastive loss can be formulated as follows:

\begin{equation}
\label{improved_loss}
\mathcal{L} = -\frac{1}{M}\sum_{m=1}^M log \Bigg [\frac{e^{{f^{m}_{h}}\cdot f^{m}_{s}} +  e^{1 - {f^{m}_{s}}\cdot \overline{f}^{m}_{l}} }{e^{{f^{m}_{h}}\cdot f^{m}_{s}} + e^{1 - {f^{m}_{s}}\cdot \overline{f}^{m}_{l}} + \sum_{j \neq m}^M e^{{f^{m}_{h}}\cdot f^{j}_{s}}} \Bigg ]
\end{equation}

While minimizing this loss, {$f^{m}_{s}$} will be pushed away by {$\overline{f}^{m}_{l}$} and pulled towards {$f^{m}_{h}$} simultaneously which minimizes the distance between the positive pair more effectively.

%In addition, it is beneficial to adaptively focus on the distance of pair ({$\overline{f}^m_{l}$}, {$f^{m}_{s}$}) and ({$f^{m}_{s}$}, {$f^{m}_{h}$}) as each individual {$\overline{f}^{m}_{l}$} loses different amounts of high-frequency details. Since the amount of the lost high-frequency details is positively related to the distance between {$\overline{f}^m_{l}$} and {$f^{m}_{h}$}, we exploit that distance to compute the penalty {$p_{m}$} and assign it to the corresponding pairs ({$\overline{f}^m_{l}$}, {$f^{m}_{s}$}) and ({$f^{m}_{s}$}, {$f^{m}_{h}$}) as shown in Figure~\ref{ccloss}b. The penalty is defined by:
Second, we exploit the condition $\overline{f}^m_{l}$ to adaptively adjust the pushing force between $\overline{f}^m_{l}$ and $f^{m}_{s}$ and the pulling force between $f^{m}_{s}$ and $f^{m}_{h}$ for each $\overline{f}^m_l$. The idea is that each individual $\overline{f}^{m}_{l}$ loses different amounts of high-frequency details, and the pushing and pulling forces should be adjusted accordingly. Since the amount of lost high-frequency details positively correlates with the distance between $\overline{f}^m_{l}$ and $f^{m}_{h}$, we exploit this distance to derive a penalty factor $p_{m}$ and apply it to the corresponding pushing and pulling forces as illustrated in Fig.~\ref{ccloss}b. The penalty term is defined by:

\begin{equation}
\label{pm}
p_{m} = h({f^{m}_{h}}\cdot {\overline{f}^m_{l}} ) = 
1 - {f^{m}_{h}} \cdot {\overline{f}^m_{l}}
\end{equation}

%Based on the above idea, we add the penalty to the Eq.~\ref{improved_loss} to design our conditional contrastive loss as follows:
By incorporating this penalty term, the proposed conditional contrastive loss can be formulated by:

\begin{equation}
\label{EqCCloss}
\mathcal{L}_{SRNCE} = -\frac{1}{M}\sum_{m=1}^M log \Bigg [\frac{e^{{f^{m}_{h}}\cdot f^{m}_{s} - p_{m}} +  e^{1 - {f^{m}_{s}}\cdot \overline{f}^{m}_{l} - p_{m}} }{e^{{f^{m}_{h}}\cdot f^{m}_{s} - p_{m}} + e^{1 - {f^{m}_{s}}\cdot \overline{f}^{m}_{l} - p_{m}} + \sum_{j \neq m}^M e^{{f^{m}_{h}}\cdot f^{j}_{s}}} \Bigg ]
\end{equation}

%While minimizing the conditional contrastive loss, the higher penalty will enhance the compactness of ({$f^{m}_{h}$}, {$f^{m}_{s}$}) and the discrepancy of ({$f^{m}_{s}$}, {$\overline{f}^m_{l}$}).
While minimizing the proposed conditional contrastive loss, the penalty term will enhance the pulling force between {$f^{m}_{h}$} and {$f^{m}_{s}$} as well as the pushing force between $f^{m}_{s}$ and $\overline{f}^m_{l}$ simultaneously.

%According to Eq.~\ref{EqCCloss} and Eq.~\ref{pm}, if the distance between ({$f^{m}_{h}$}, {$\overline{f}^m_{l}$}) is larger, its corresponding pairs ({$f^{m}_{h}$}, {$f^{m}_{s}$}) and ({$\overline{f}^m_{l}$}, {$f^{m}_{s}$}) will be assigned larger penalties.  

Besides the conditional contrastive loss, we introduce a HR reconstruction loss {$\mathcal{L}^{hr}_{rec}$} to ensure that {$f_{h}$} capture most high-frequency features:

\begin{equation}
\mathcal{L}_{rec}^{hr} =  \Arrowvert I_{h} - D_{rec}(f_{h}) \Arrowvert_1{}
\end{equation}

Together with the SR reconstruction loss {$\mathcal{L}^{sr}_{rec} = \Arrowvert I_{h} - I_{s} \Arrowvert$}  (a L1 loss), the overall objective function for the contrastive feature refinement can be defined as follows:
\begin{equation}
\mathcal{L}_{CFRS} = \mathcal{L}_{SRNCE} + \mathcal{L}^{sr}_{rec} + \mathcal{L}_{rec}^{hr}
\end{equation}

\begin{table*}[!htbp]%[tbp]%
%   \scriptsize
  \small
  \caption{Ablation study of contrastive decoupling encoding and conditional contrastive loss in CRL-SR. The evaluations were performed over dataset Set14 with a scaling factor of $\times$4, a spatially variant kernel width from 0.2 to 4 and a noise level from 5 to 50.}
  \renewcommand\tabcolsep{9pt}
  \label{ablation}
  \centering
  \begin{center}
      \begin{tabular}{l|c|c|c|c|c|c}
        \hline
        Models & a & b & c & d & e & f\\
        \hline
        \hline
        Contrastive Decoupling Encoding (Ours)& $\times$ & $\times$ & $\times$ & $\checkmark$ & $\checkmark$ & $\checkmark$  \\
        % Condition-based Weighted Contrastive Loss (Ours)& $\times$ & $\checkmark$ & $\times$ & $\times$ & $\times$ & $\checkmark$  \\
        Conditional Contrastive Loss (Ours)& $\times$ & $\checkmark$ & $\times$ & $\times$ & $\times$ & $\checkmark$  \\
        InfoNCE Loss \cite{oord2018representation} & $\times$ & $\times$ & $\checkmark$ & $\times$ & $\checkmark$ & $\times$  \\
        \hline
        PSNR on Set14\cite{zeyde2010single} (4$\times$) & 23.62 & 24.11 & 23.87 & 24.08 & 24.14 & 24.31 \\
        \hline
      \end{tabular}
    %   }
  \end{center}
  \vspace{-3mm}
\end{table*}

\section{Experiments}
\subsection{Datasets and Training Settings}
\label{training_details}
Following \cite{wang2021unsupervised, xu2020unified, gu2019blind}, we train our network by using the training sets of DIV2K \cite{agustsson2017ntire} and Flickr2K\cite{timofte2017ntire}. The evaluations (in PSNR) are performed over four standard datasets Set5 \cite{bevilacqua2012low}, Set14 \cite{zeyde2010single}, B100 \cite{martin2001database} and Urban100 \cite{huang2015single}. We apply the degradation in Eq.\ref{eq1} to generate LR images in both training and testing. Specifically, we first train a model by applying degradations of isotropic Gaussian kernels and noises, where the kernel size is fixed at {$21 \times 21$}, the kernel width is set to the range{$[0.2, 4.0]$} and the noise level is set at {$[0, 75]$} as in \cite{xu2020unified, wang2021unsupervised}. We also train our model on degradations with anisotropic Gaussian kernels and noises, where the kernels have a Gaussian density function {$N(0,\Sigma)$}, the covariance matrix {$\Sigma$} is determined by a random rotation angle {$\theta\sim{U(0,\pi)}$} and two random eigenvalues {$\lambda_1,\lambda_2 \sim{U(0.2,4)}$}, and the noise is set to the range {$[0, 25]$}, as in \cite{wang2021unsupervised}

In each training batch, 32 LR patches of size {$48 \times 48$} are extracted as inputs. Similar to \cite{gu2019blind, wang2021unsupervised, zhang2018residual}, we augment the training images by random rotation by $90^\circ$, $180^\circ$, $270^\circ$ as well as horizontal flipping. The SR is evaluated over the Y channel in the YCbCr space. We adopt Adam optimizer with {$\beta_{1}$} = 0.9, {$\beta_{2}$} = 0.999, and $\epsilon$ = {$10^{-8}$}. The initial learning rate is set to 0.0001 and then halved every 200 epochs. In the experiments, we fix parameter {$\tau$} in the contrastive loss to 0.07 as in \cite{chen2020simple}. The dimension \textit{S} of the feature vectors after the MLP mapping is 256. Our baseline only contains an encoder (15 residual blocks) for feature extraction and a decoder for upsampling LR images. We use the Pytorch framework to implement our model with NVIDIA Titan V.

\begin{table*}[tbp]%[!htbp]
  \scriptsize
  \caption{%Average PSNR values on spatially variant degradations. The results highlighted in {\color{gray}gray} color indicates the unfair comparison since DASR has deeper baseline which is based on residual groups.
  SR performance under the presence of spatially variant degradation (in PSNR): The experiments involve degradation distributions that have spatially variant kernel width and noise levels. Note DASR employs a much deeper network architecture with residual groups and degradation-aware blocks.}
  \label{spatially variations}
  \vspace{5pt}
  \centering
  \setlength{\tabcolsep}{1.12mm}{
  \begin{tabular}{|l|c|c|ccc|ccc|ccc|ccc|}
    \hline
    \multirow{2}{*}{Methods} & \multirow{2}{*}{Kernel width} & \multirow{2}{*}{Noise level} & \multicolumn{3}{c|}{\multirow{1}{*}{Set5\cite{bevilacqua2012low}}} &
    \multicolumn{3}{c|}{\multirow{1}{*}{Set14\cite{zeyde2010single}}}& \multicolumn{3}{c|}{\multirow{1}{*}{B100\cite{martin2001database}}} & \multicolumn{3}{c|}{\multirow{1}{*}{Urban100\cite{huang2015single}}}\\
    & & & $\times$2 & $\times$3 & $\times$4 & $\times$2 & $\times$3 & $\times$4 & $\times$2 & $\times$3 & $\times$4 & $\times$2 & $\times$3 & $\times$4\\ 
    \hline
    \hline
    RCAN~\cite{zhang2018image} & \multirow{4}{*}{[0.2, 4]} & \multirow{4}{*}{5} & 26.57 & 26.64 & 26.45 & 25.13 & 25.04 & 24.80 & 25.33 & 25.20 & 24.96 & 22.68 & 22.59 & 22.40 \\
    IKC~\cite{gu2019blind} & & & 29.15 & 28.03 & 27.44 & 27.35 & 26.19 & 25.57 & 26.83 & 25.94 & 25.45 & 24.63 & 23.72 & 22.94 \\
    DASR~\cite{wang2021unsupervised} & & & {\color{gray}30.12} & {\color{gray}28.89} & {\color{gray}28.18} & {\color{gray}27.92} & {\color{gray}26.81} & {\color{gray}26.24} & {\color{gray}27.19} & {\color{gray}26.38} & {\color{gray}25.89} & {\color{gray}25.12} & {\color{gray}24.18} & {\color{gray}23.53}  \\
    CRL-SR (Ours) & & & \textbf{30.31} & \textbf{29.50} & \textbf{28.66} & \textbf{28.01} & \textbf{27.19} & \textbf{26.40} & \textbf{27.33} & \textbf{26.65} & \textbf{26.00} & \textbf{25.25} & \textbf{24.50} & \textbf{23.76}\\ 
    \hline
    RCAN~\cite{zhang2018image} & \multirow{4}{*}{2} & \multirow{4}{*}{[5, 50]} & 19.83 & 19.48& 19.64 & 19.44 & 19.08 & 19.16 & 19.42 & 19.09 & 19.17 & 18.61 & 18.27 & 18.27 \\
    IKC~\cite{gu2019blind} & & & 26.63 & 25.68 & 24.95 & 25.12 & 24.41 & 23.85 & 25.05 & 24.39 & 23.88 & 23.03 & 22.40 & 21.78 \\
    DASR~\cite{wang2021unsupervised} & & & {\color{gray}27.59} & {\color{gray}26.52} & {\color{gray}25.58} & {\color{gray}25.75} & {\color{gray}25.06} & {\color{gray}24.40} & {\color{gray}25.44} & {\color{gray}24.82} & {\color{gray}24.27} & {\color{gray}23.59} & {\color{gray}22.88} & {\color{gray}22.24} \\
    CRL-SR (Ours) & & & \textbf{27.84} & \textbf{26.73} & \textbf{25.98} & \textbf{25.91} & \textbf{25.22} & \textbf{24.61} & \textbf{25.59} & \textbf{24.97} & \textbf{24.43} & \textbf{23.71} & \textbf{23.04} & \textbf{22.44}\\ 
    \hline
    RCAN~\cite{zhang2018image} & \multirow{4}{*}{[0.2, 4]} & \multirow{4}{*}{[5, 50]} & 19.69 & 19.32 & 19.46&19.36 & 18.95 & 19.01 & 19.42 & 19.03 & 19.09 & 18.52 & 18.19 & 18.15\\
    IKC~\cite{gu2019blind} & & & 25.92 & 24.96 & 24.52 & 24.68 & 23.97 & 23.40 & 24.88 & 24.21 & 23.76 & 22.45 & 21.92 & 21.36 \\
    DASR~\cite{wang2021unsupervised} & & & {\color{gray}26.61} & {\color{gray}25.78} & {\color{gray}25.11} & {\color{gray}25.28} & {\color{gray}24.59} & {\color{gray}24.05} & {\color{gray}25.27} & {\color{gray}24.64} & {\color{gray}24.12} & {\color{gray}22.98} & {\color{gray}22.44} & {\color{gray}21.88} \\
    CRL-SR (Ours) & & & \textbf{27.54} & \textbf{26.55} & \textbf{25.67} & \textbf{25.84} & \textbf{25.06} & \textbf{24.44} & \textbf{25.65} & \textbf{24.92} & \textbf{24.35} & \textbf{23.56} & \textbf{22.83} & \textbf{22.22}\\ 
    \hline
  \end{tabular}}
  \vspace{-17pt}
\end{table*}

\subsection{Ablation Study}
\label{ablation_study}
To examine the effectiveness of the proposed contrastive decoupling encoding and contrastive feature refinement, we perform ablation experiments to study how they affect the SR in PSNR. In the ablation studies, we adopt a small network architecture that has only a few convolutional layers and residual blocks in its encoders and decoders (for all the models \textit{a-f} in Table \ref{ablation}) which allows to demonstrate the effects of network and loss designs more clearly. The model \emph{$a$} works as a baseline which contains the encoder \emph{$\overline{E}_{LR}$} and the decoder \emph{$D_{ref}$} only. As Table \ref{ablation} shows, the baseline achieves a PSNR of 23.62 dB on the dataset Set14 with the adopted small network (More details about the network architecture are provided in supplementary material).

We replace $\overline{E}_{LR}$ of model $a$ by CDE which improves the PSNR from 23.62 dB to 24.08 dB. We also include CFR with the conditional contrastive loss into the model $a$ which improves the PSNR from 23.62 dB to 24.11 dB. Both studies demonstrate the effectiveness of the proposed CDE and CFR. In addition, we compare the conditional contrastive loss with InfoNCE which improves PSNR from 23.87dB to 24.11dB. The better performance is largely because the proposed conditional contrastive loss takes \emph{$\overline{F}_{LR}$} as a condition which introduces an extra pushing force and adjusts the pushing and pulling forces adaptively. Finally, the complete CRL-SR with both CDE and CFR performs the best with a PSNR of 24.31 dB, showing the complementarity of the proposed CDE and CFR.

\subsection{Experiments on Spatially Variant Degradations}

The proposed CRL-SR can effectively handle spatially variant degradation as widely observed in real-world data. We evaluate it over the four synthetic sets where LR images are synthesized with spatially variant blur kernels and noises with kernel width and noise level gradually increasing in the range of [0.2, 4] and [5, 50] from the left to the right of images as in \cite{xu2020unified}. We compare CRL-SR with three representative SR methods. The first is state-of-the-art non-blind SR method RCAN which works best for certain known bicubic downsampling. The other two are state-of-the-art blind SR methods IKC and DASR, where DASR adopts a deeper network architecture with residual groups and degradation-aware blocks. We re-trained IKC and DASR on isotropic Gaussian kernels and noises since they are pre-trained with degradation with isotropic Gaussian kernels only. 

\begin{figure*}[tbp]%[!htbp]
\begin{center}
\includegraphics[width=1\textwidth]{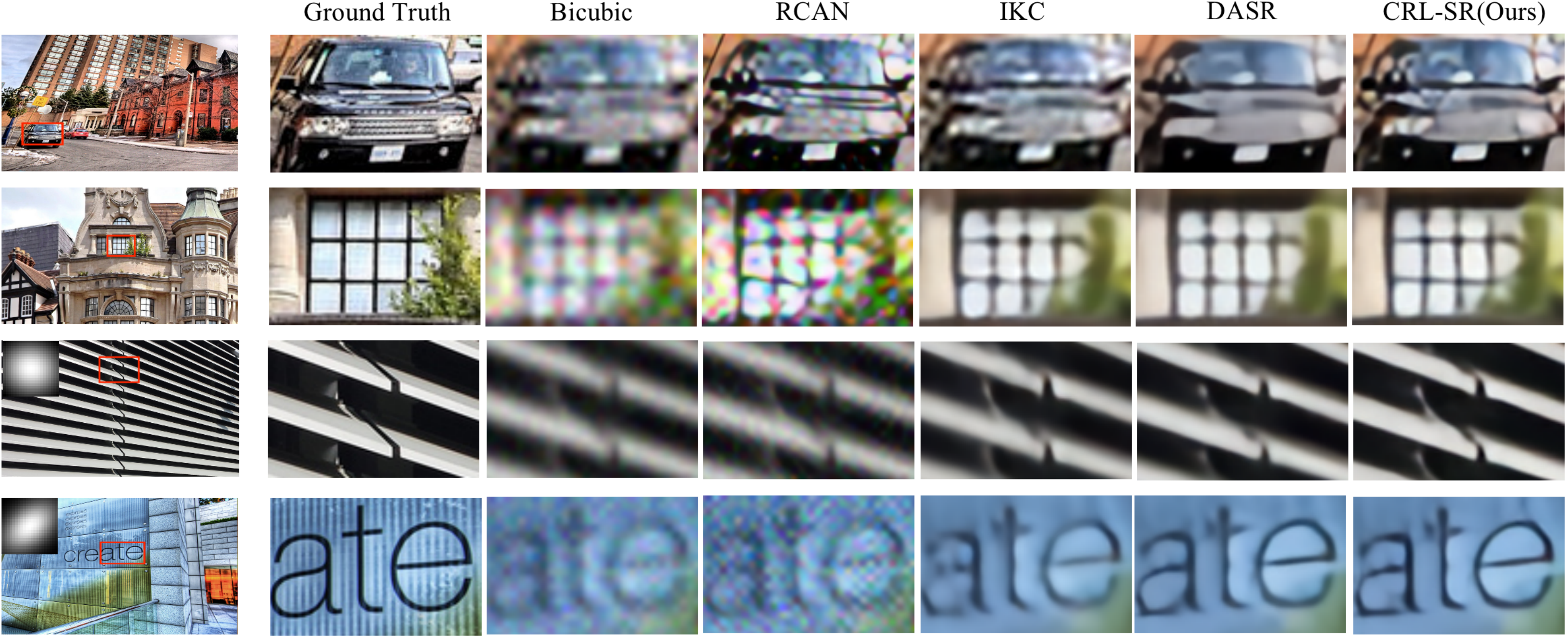}
\end{center}
\caption{Qualitative experiments on the dataset Urban100 with a scale factor 4: The two sample images in the first two rows are degraded by spatially variant blurs and noises, while the two in the last two rows suffer from multi-modal degradation.%Visual comparisons on the Urban100 dataset with scale factor 4. The first two rows show the results on the spatial variant blur and noise degradation. The last two rows show the results on the multi-modal degradation.
}
\label{visual-comparison}
\end{figure*}

As Table \ref{spatially variations} shows, RCAN suffers from clear performance drops when the degradation is beyond bicubic downsampling. IKC struggles in explicit blur kernel estimation when the degradation is multi-modal and spatially variant. Although DASR has deeper and more powerful architecture and introduces implicit degradation representation to deal with multi-modal degradation, it still cannot handle spatially variant degradation well. As a comparison, CRL-SR performs clearly better than the three methods and it outperforms by large margins when both blur kernels and noise levels are spatially variant (more challenging to SR task). The proposed CRL-SR is less affected because the proposed contrastive decoupling encoding and contrastive feature refinement directly disentangle low-frequency features (instead of estimating degradation) and generate lost high-frequency details . The experiment results are well aligned with the illustration in Fig.~\ref{visual-comparison} where CRL-SR produces high-fidelity SR with more details and less artifacts.

\begin{table*}[tbp]%[!htbp]
    \caption{%Comparison with state-of-the-art PSNR-driven SR methods on the generalization of isotropic Gaussian kernels and the noise levels, that are outside training data. Average PSNR results achieved on Set5\cite{bevilacqua2012low}. The results highlighted in {\color{gray}gray} color indicates the unfair comparison since DASR has deeper baseline and better architecture.
    SR performance for the generalization in isotropic Gaussian kernels and the noise levels which are unseen in training data. The experiments were conducted on dataset Set5\cite{bevilacqua2012low} with evaluation metric PSNR. Note DASR employs a much deeper network architecture with residual groups and degradation-aware blocks.}
    \label{generalize_iso}
    \begin{center}
      \scriptsize
      \setlength{\tabcolsep}{1.00mm}{
        \begin{tabular}{|l|c|ccc|ccc|ccc|ccc|ccc|}
          \hline 
          \multirow{2}{*}{Method} & \multirow{2}{*}{\tabincell{c}{Scale}} & \multicolumn{15}{c|}{\multirow{1}{*}{Kernel Width}}
          \tabularnewline
          &
          & \multicolumn{3}{c}{\multirow{1}{*}{$\varepsilon$ = 4.0}} 
          & \multicolumn{3}{c}{\multirow{1}{*}{$\varepsilon$ = 4.5}} 
          & \multicolumn{3}{c}{\multirow{1}{*}{$\varepsilon$ = 5.0}} & \multicolumn{3}{c}{\multirow{1}{*}{$\varepsilon$ = 5.5}}   & \multicolumn{3}{c|}{\multirow{1}{*}{$\varepsilon$ = 6.0}} 
          \tabularnewline
          \hline 
          \hline
          \multicolumn{2}{|l|}{\multirow{1}{*}{Noise Level}} 
          & 90 & 95 & 100 
          & 90 & 95 & 100 
          & 90 & 95 & 100 
          & 90 & 95 & 100 
          & 90 & 95 & 100 
          \tabularnewline
          \hline
          RCAN \cite{zhang2018image} & \multirow{4}{*}{$\times2$}
          & 12.82 & 12.51 & 12.37
          & 12.78 & 12.46 & 12.30
          & 12.74 & 12.38 & 12.35
          & 12.68 & 12.34 & 12.32
          & 12.57 & 12.29 & 12.33
          \tabularnewline
          IKC \cite{gu2019blind}
          & 
          & 22.93 & 21.62 & 21.33
          & 21.58 & 21.32 & 21.05
          & 21.30 & 21.15 & 20.83
          & 21.12 & 20.96 & 20.65
          & 21.95 & 20.74 & 20.51
          \tabularnewline
          DASR \cite{wang2021unsupervised}
          &
          & {\color{gray}23.06} & {\color{gray}22.79} & {\color{gray}22.47}
          & {\color{gray}22.73} & {\color{gray}22.47} & {\color{gray}22.18}
          & {\color{gray}22.46} & {\color{gray}22.22} & {\color{gray}21.95}
          & {\color{gray}22.24} & {\color{gray}22.02} & {\color{gray}21.76}
          & {\color{gray}22.07} & {\color{gray}21.86} & {\color{gray}21.62}
          \tabularnewline
          CRL-SR (Ours)  
          &
          & \textbf{23.21} & \textbf{23.07} & \textbf{22.92} 
          & \textbf{22.90} & \textbf{22.76} & \textbf{22.63} 
          & \textbf{22.63} & \textbf{22.51} & \textbf{22.39} 
          & \textbf{22.42} & \textbf{22.31} & \textbf{22.19} 
          & \textbf{22.25} & \textbf{22.14} & \textbf{22.03}
          \tabularnewline
          \hline
          \hline
          RCAN \cite{zhang2018image} & \multirow{4}{*}{$\times3$}
          & 12.67 & 12.49 & 12.32
          & 12.62 & 12.44 & 12.27
          & 12.57 & 12.40 & 12.23
          & 12.54 & 12.36 & 12.20
          & 12.51 & 12.34 & 12.18
          \tabularnewline
          IKC \cite{gu2019blind}
          & 
          & 21.10 & 20.73 & 20.35
          & 20.84 & 20.44 & 20.12
          & 20.62 & 20.30 & 19.98
          & 20.45 & 20.14 & 19.84
          & 20.33 & 20.01 & 19.68
          \tabularnewline
          DASR \cite{wang2021unsupervised}
          &
          & {\color{gray}22.21} & {\color{gray}21.87} & {\color{gray}21.47}
          & {\color{gray}21.97} & {\color{gray}21.63} & {\color{gray}21.25} 
          & {\color{gray}21.75} & {\color{gray}21.43} & {\color{gray}21.06}
          & {\color{gray}21.58} & {\color{gray}21.27} & {\color{gray}20.92}
          & {\color{gray}21.45} & {\color{gray}21.15} & {\color{gray}20.80} 
          \tabularnewline
          CRL-SR (Ours)  
          &
          & \textbf{22.52} & \textbf{22.36} & \textbf{22.20} 
          & \textbf{22.29} & \textbf{22.15} & \textbf{22.00} 
          & \textbf{22.10} & \textbf{21.97} & \textbf{21.83} 
          & \textbf{21.95} & \textbf{21.82} & \textbf{21.68} 
          & \textbf{21.81} & \textbf{21.69} & \textbf{21.57}
          \tabularnewline       
          \hline
          \hline
          RCAN \cite{zhang2018image} & \multirow{4}{*}{$\times4$}
          & 13.36 & 13.17 & 13.00
          & 13.30 & 13.12 & 12.95
          & 13.26 & 13.07 & 12.90
          & 13.22 & 13.03 & 12.87
          & 13.19 & 13.00 & 12.84
          \tabularnewline
          IKC \cite{gu2019blind}
          & 
          & 20.56 & 20.38 & 20.06
          & 20.32 & 20.15 & 19.93
          & 20.22 & 20.03 & 19.75
          & 20.04 & 19.88 & 19.63
          & 19.97 & 19.74 & 19.55
          \tabularnewline
          DASR \cite{wang2021unsupervised}
          &
          & {\color{gray}21.73} & {\color{gray}21.51} & {\color{gray}21.28}
          & {\color{gray}21.55} & {\color{gray}21.34} & {\color{gray}21.11}
          & {\color{gray}21.40} & {\color{gray}21.20} & {\color{gray}20.98}
          & {\color{gray}21.27} & {\color{gray}21.08} & {\color{gray}20.87}
          & {\color{gray}21.17} & {\color{gray}20.98} & {\color{gray}20.78}
          \tabularnewline
          CRL-SR (Ours)  
          &
          & \textbf{21.90} & \textbf{21.75} & \textbf{21.61} 
          & \textbf{21.73} & \textbf{21.59} & \textbf{21.44} 
          & \textbf{21.58} & \textbf{21.44} & \textbf{21.31} 
          & \textbf{21.45} & \textbf{21.32} & \textbf{21.19} 
          & \textbf{21.34} & \textbf{21.22} & \textbf{21.10}
          \tabularnewline 
          \hline  
      \end{tabular}
      }
    \end{center}
    \vspace{-0.2cm}
\end{table*}

\subsection{Experiments on Multi-Modal Degradations.}

%In addition to conduct experiments on degradations with spatial variations, we also evaluate our proposed CRL-SR on degradations with multi-modal distributions to prove the effectiveness of our model. In these evaluations, we utilize two types of degradation and the details of the experiments are shown in the following paragraphs.

%We also evaluated CRL-SR when the degradation has a multi-modal distribution. Specifically, we tested with two types of degradation as to be described in the ensuing two subsections.

%To ensure that CRL-SR is more suitable for super-resolving real-world images, we also evaluate the generalization of CRL-SR when handling the multi-modal degradations. Specifically, we test with two types of degradation as to be described in the ensuing two subsections.
To examine how CRL-SR performs for real-world images, we evaluate it while LR images contain two types of  multi-modal degradation that are unseen during the training stage.

\paragraph{Degradation with Isotropic Gaussian Kernels and Noises. }

%The first type of degradation consists of bicubic downsampling, isotropic Gaussian kernels and noises, where isotropic Gaussian kernels generate Gaussian blurs. Specifically, we employ three scale factors 2, 3, and 4 for bicubic downsampling. The kernel width of the isotropic kernels is set at 0.2, 1.3 and 2.6, and the noise level is set at 15 and 50. As Table \ref{multiple} shows, we compare CRL-SR with RDN, RCAN and IKC over four benchmark datasets Set5, Set14, B100 and Urban100. Similarly, RDN and RCAN produce poor SR performance as the degradation has a multi-modal distribution. In addition, CRL-SR outperforms IKC consistently with similar reasons.

The first type of multi-modal degradation consists of bicubic downsampling, isotropic Gaussian kernels and noises, where isotropic Gaussian kernels generate Gaussian blurs. We employ three scale factors 2, 3, and 4 for bicubic downsampling. The width of the isotropic kernel is set at 4.0, 4.5, 5.0, 5.5 and 6.0, and the noise level is set at 90, 95 and 100. Note all kernel widths and noise levels do not appear in the training stage. Table \ref{generalize_iso} shows experimental results on the dataset Set5. It can be observed that CRL-SR generalizes clearly better for blurred and noisy images due to our innovative designs that exploit contrastive learning to directly remove degradation and generate lost high-frequency details. In addition, CRL-SR outperforms DASR clearly, demonstrating that the proposed CDE and CFR can work with simple network architectures effectively.

\paragraph{Degradations with Anisotropic Gaussian Kernels and Noises. } 
%The second type of degradation consists of bicubic downsampling, anisotropic Gaussian kernels and noises, where the anisotropic Gaussian kernel is equivalent to the combination of isotropic Gaussian kernel and motion blur. We use 9 typical blur kernels and three noise levels in evaluations as shown in Table ~\ref{tab2}. Since RDN and RCAN consider bicubic downsampling only, they do not perform well with multiple degradations. In addition, IKC works better but still suffers from clear kernel estimation errors while handling complex degradations. As a comparison, the proposed CRL-SR performs clearly better as it removes disentangle resolution-invariant features and reconstructs high-frequency details effectively. In addition, CRL-SR outperforms the three methods with larger margins when the kernel size increases, largely because larger kernel size makes SR reconstruction more challenging but it does not affect CRL-SR much due to the proposed contrastive decoupling encoding and contrastive feature refinement.

The second type of degradation consists of bicubic downsampling, anisotropic Gaussian kernels and noises, where the anisotropic Gaussian kernel is equivalent to the combination of isotropic Gaussian kernel and motion blur. We use 9 typical blur kernels and three noise levels in evaluations as shown in Table ~\ref{tab2}. Note that at least one eigenvalue for each blur kernel exceeds the range set in training.
We retrained IKC for fair comparison as IKC is not pre-trained on anisotropic Gaussian kernel and noises. Since RCAN considers bicubic downsampling only, it does not generalize to multi-modal degradation well. IKC works better but still suffers from clear kernel estimation errors while handling multi-modal degradation. As a comparison, CRL-SR performs clearly better, and it even outperforms DASR due to the proposed CDE and CFR. In addition, CRL-SR outperforms with larger margins for similar reasons when the kernel size increases. As shown in Fig.~\ref{visual-comparison}, CRL-SR produces high-fidelity SR with sharper edges in texts under the presence of unseen multi-modal degradation.

\subsection{Experiments on Real-World Images}

\begin{figure*}[tbp]%[!htbp]
\begin{center}
\includegraphics[width=1\textwidth]{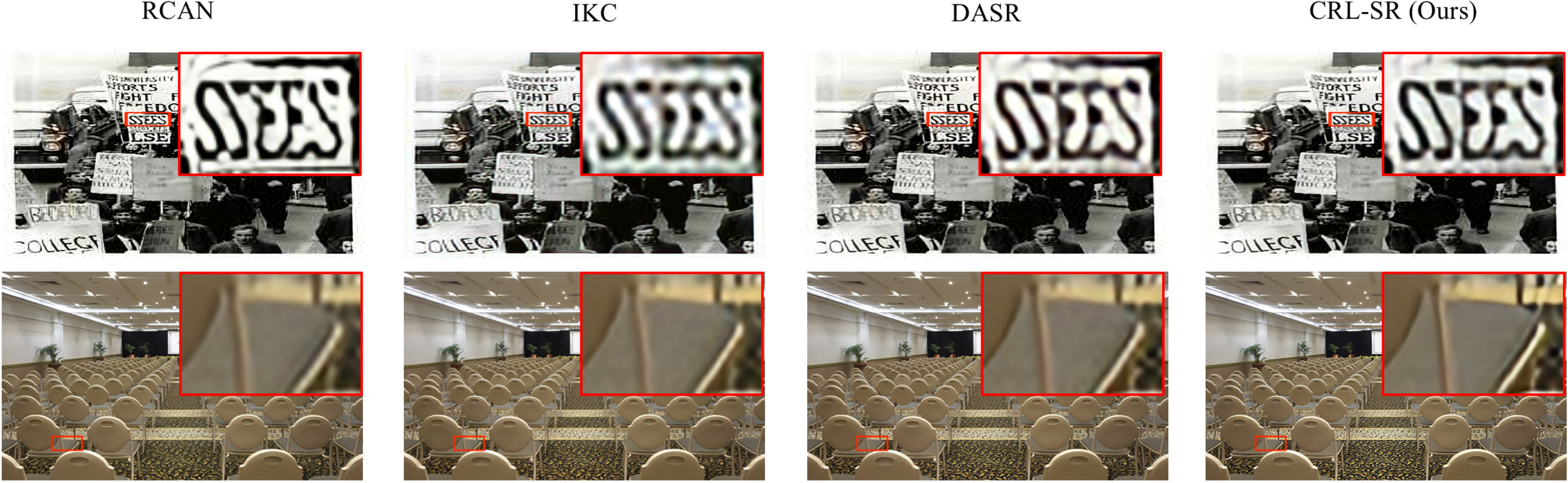}
\end{center}
\caption{%The visual comparison on the real-world images with scale factor 4.
Qualitative experiments on real-world images with a scale factor 4: The proposed CRL-SR successfully reproduces faithful image edges and details, demonstrating its capability in handling spatially variant, multi-modal and unknown degradation.
}
\label{real-comparison}
\end{figure*}

We also evaluated CRL-SR with real-world images. Following prior studies in~\cite{zhang2018learning, zhang2018image, zhang2018residual, wang2021unsupervised, gu2019blind}, we just qualitatively evaluate CRL-SR and compare it with three state-of-the-art SR methods. As Fig.~\ref{real-comparison} shows, the proposed CRL-SR can successfully remove blurs and produce sharper edges and clear details for the texts and chairs in scenes. The evaluations on real-world images further demonstrate the effectiveness of the proposed CRL-SR while handling real-world degradation which often has spatially variant, multi-modal and unknown distributions.
%We further extend experiments to real images. FolloWe further extend experiments to real images. Following most previous methods \cite{zhang2018learning, zhang2018image, zhang2018residual, wang2021unsupervised, gu2019blind}, we use visual results to compare the performance of different methods on super-resolving real images. {\color{red} Fig.~\ref{real-comparison} show the visual comparisons between our CRL-SR and other methods on some real images. We can observe that only our CRL-SR can effectively remove blur and produce clearer edges of texts and chairs.} The results show the effectiveness of our CRL-SR in removing degradations and generating high-frequency details. wing most previous methods \cite{zhang2018learning, zhang2018image, zhang2018residual, xu2020unified, gu2019blind}, we use visual results to compare the performance of different methods on super-resolving real images. {\color{red} Fig.~\ref{real-comparison} show the visual comparisons between our CRL-SR and other methods on some real images. We can observe that only our CRL-SR can effectively remove blur and produce clearer edges of texts and chairs.} The results show the effectiveness of our CRL-SR in removing degradations and generating high-frequency details.

\begin{table*}[tbp]%[!htbp]
    \caption{
    %Average PSNR values on Set14 for $\times4$ SR with noise and anisotropic Gaussian kernels. The results highlighted in {\color{gray}gray} color indicates the unfair comparison with DASR.
    SR performance under the presence of anisotropic Gaussian kernels and noises. The experiments were conducted on the dataset Set14 for $\times4$ with metric PSNR. Note DASR employs a much deeper network architecture with residual groups and degradation-aware blocks.}
    \label{tab2}
    \begin{center}
      \scriptsize
      \setlength{\tabcolsep}{1.7mm}{
        \begin{tabular}{|l|c|ccccccccc|}
          \hline 
          \multirow{2}{*}{\tabincell{c}{\\Method}}
          & \multirow{2}{*}{\tabincell{c}{\\Noise}}
          & \multicolumn{9}{c|}{Blur Kernel}
          \tabularnewline
          & 
          & \begin{minipage}[b]{0.06\columnwidth}
            \centering
            \raisebox{-.5\height}{\includegraphics[width=0.70\linewidth]{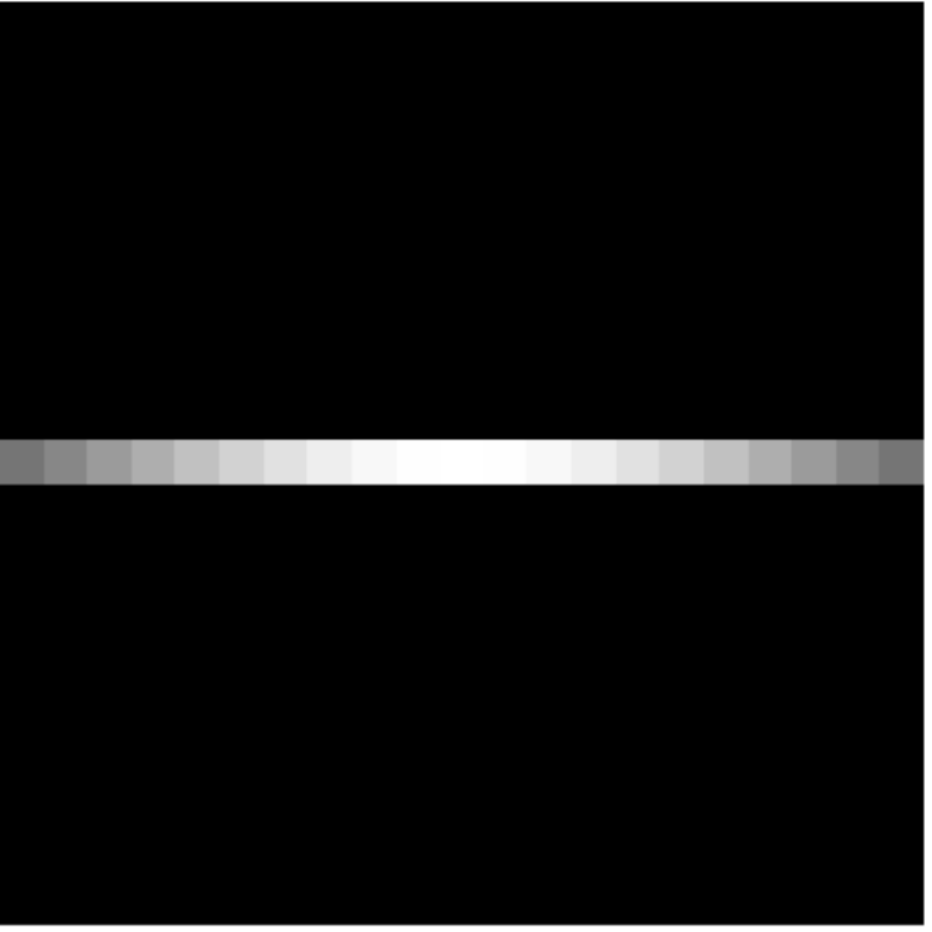}}
          \end{minipage}
          & \begin{minipage}[b]{0.06\columnwidth}
            \centering
            \raisebox{-.5\height}{\includegraphics[width=0.70\linewidth]{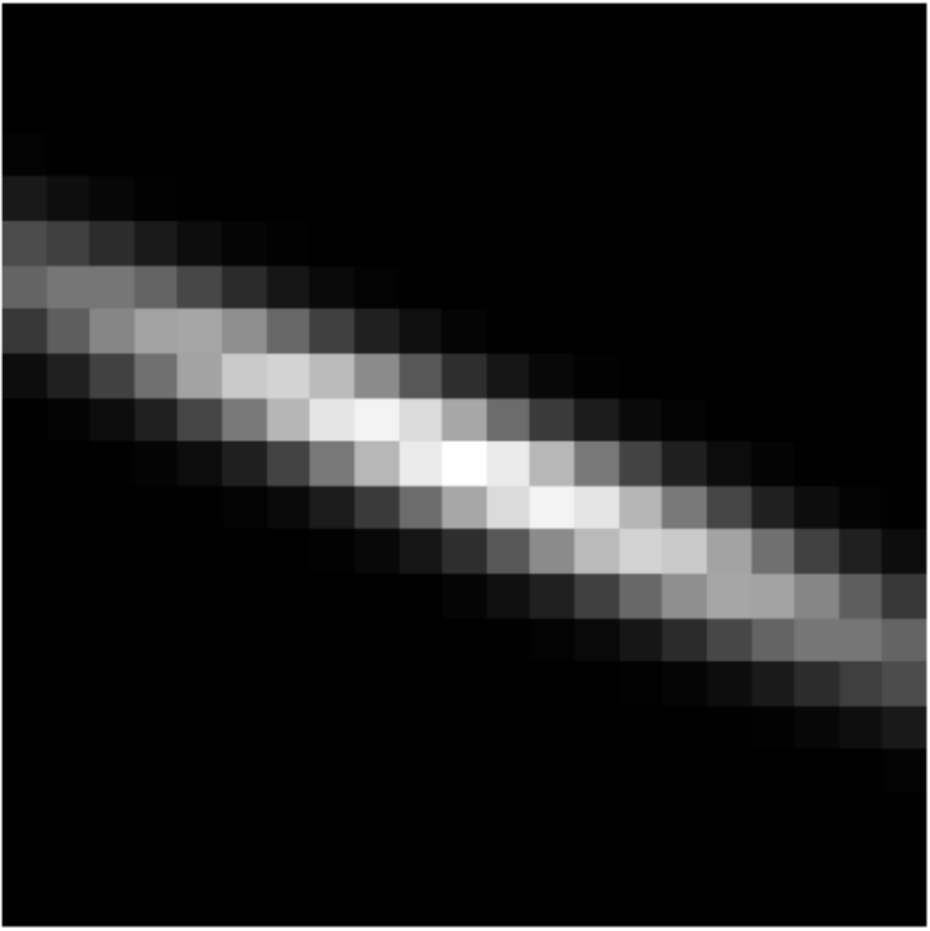}}
          \end{minipage}
          & \begin{minipage}[b]{0.06\columnwidth}
            \centering
            \raisebox{-.5\height}{\includegraphics[width=0.70\linewidth]{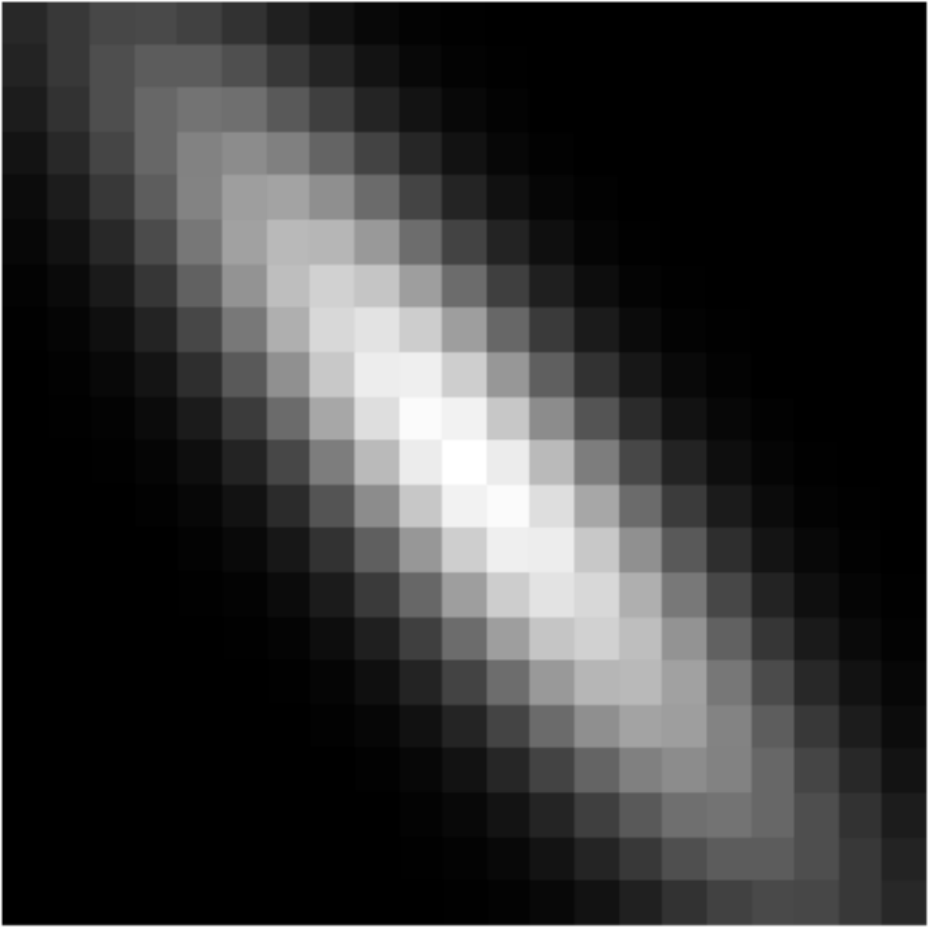}}
          \end{minipage}
          & \begin{minipage}[b]{0.06\columnwidth}
            \centering
            \raisebox{-.5\height}{\includegraphics[width=0.70\linewidth]{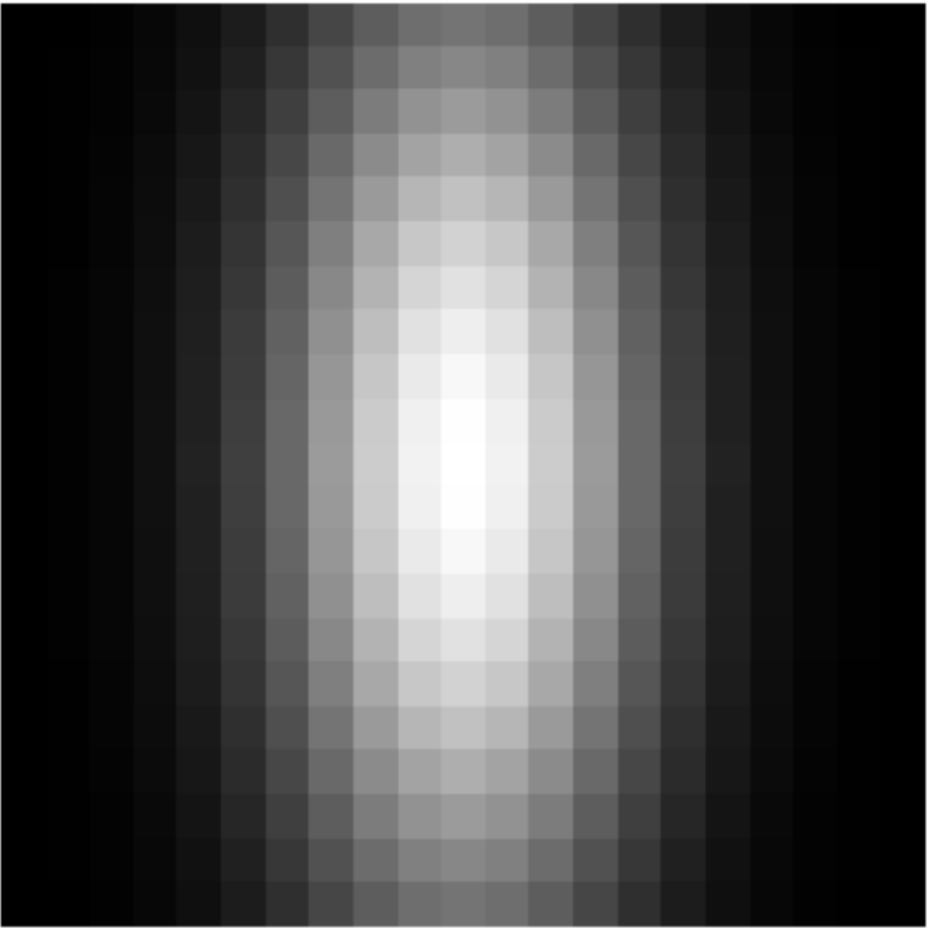}}
          \end{minipage}
          & \begin{minipage}[b]{0.06\columnwidth}
            \centering
            \raisebox{-.5\height}{\includegraphics[width=0.70\linewidth]{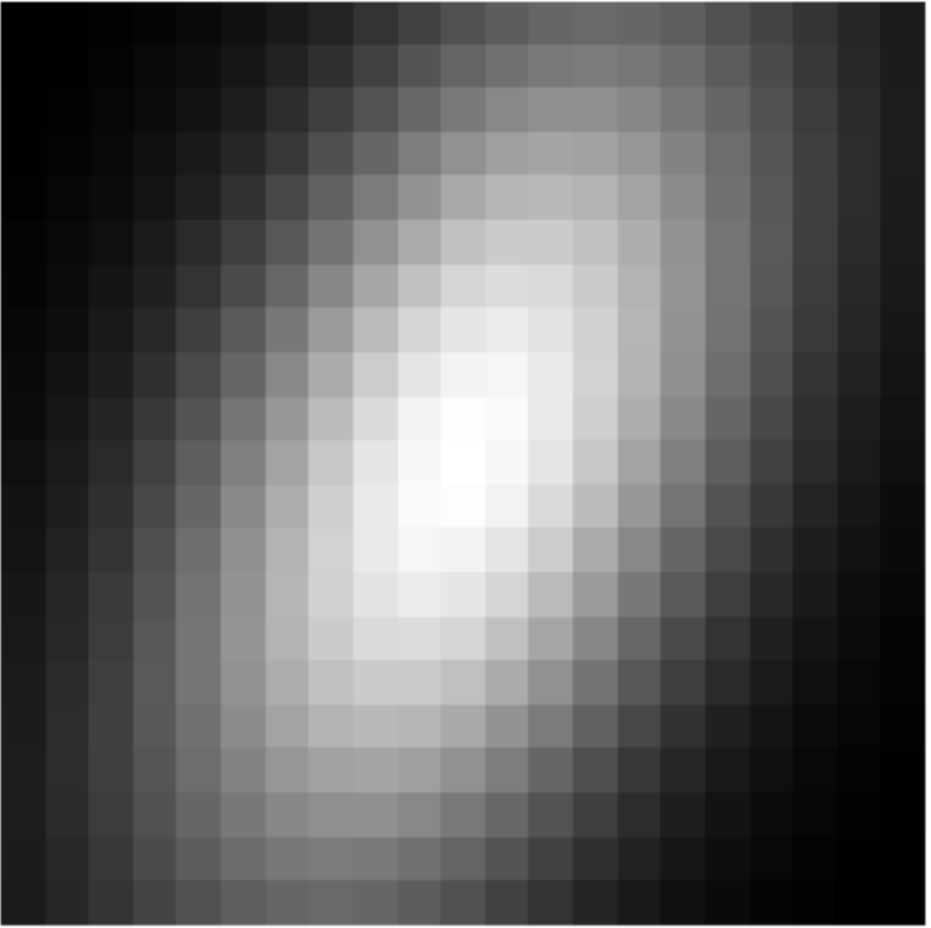}}
          \end{minipage}
          & \begin{minipage}[b]{0.06\columnwidth}
            \centering
            \raisebox{-.5\height}{\includegraphics[width=0.70\linewidth]{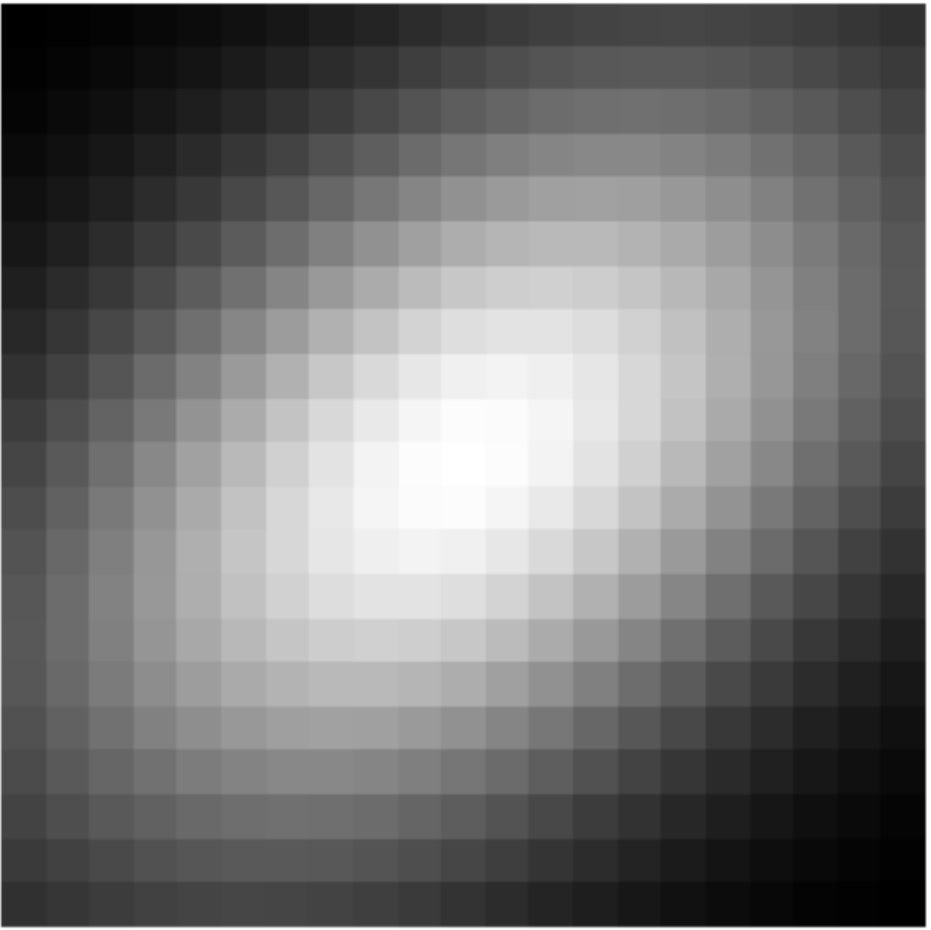}}
          \end{minipage}
          & \begin{minipage}[b]{0.06\columnwidth}
            \centering
            \raisebox{-.5\height}{\includegraphics[width=0.70\linewidth]{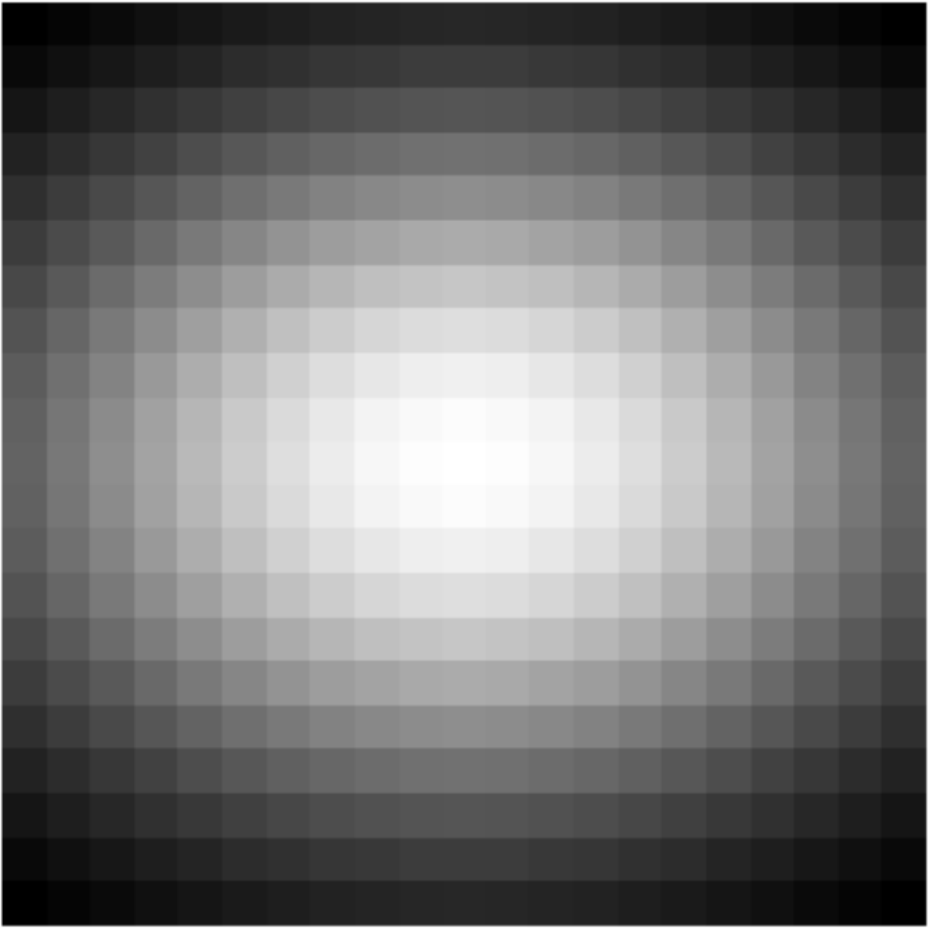}}
          \end{minipage}
          & \begin{minipage}[b]{0.06\columnwidth}
            \centering
            \raisebox{-.5\height}{\includegraphics[width=0.70\linewidth]{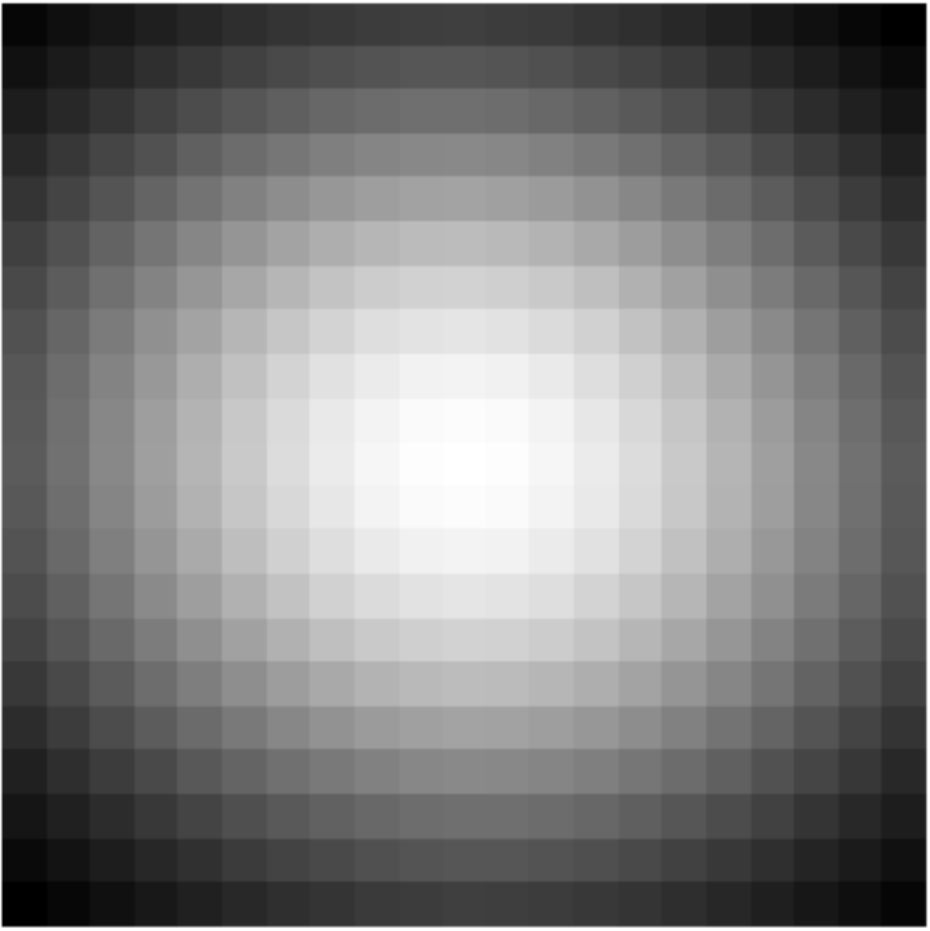}}
          \end{minipage}
          & \begin{minipage}[b]{0.06\columnwidth}
            \centering
            \raisebox{-.5\height}{\includegraphics[width=0.70\linewidth]{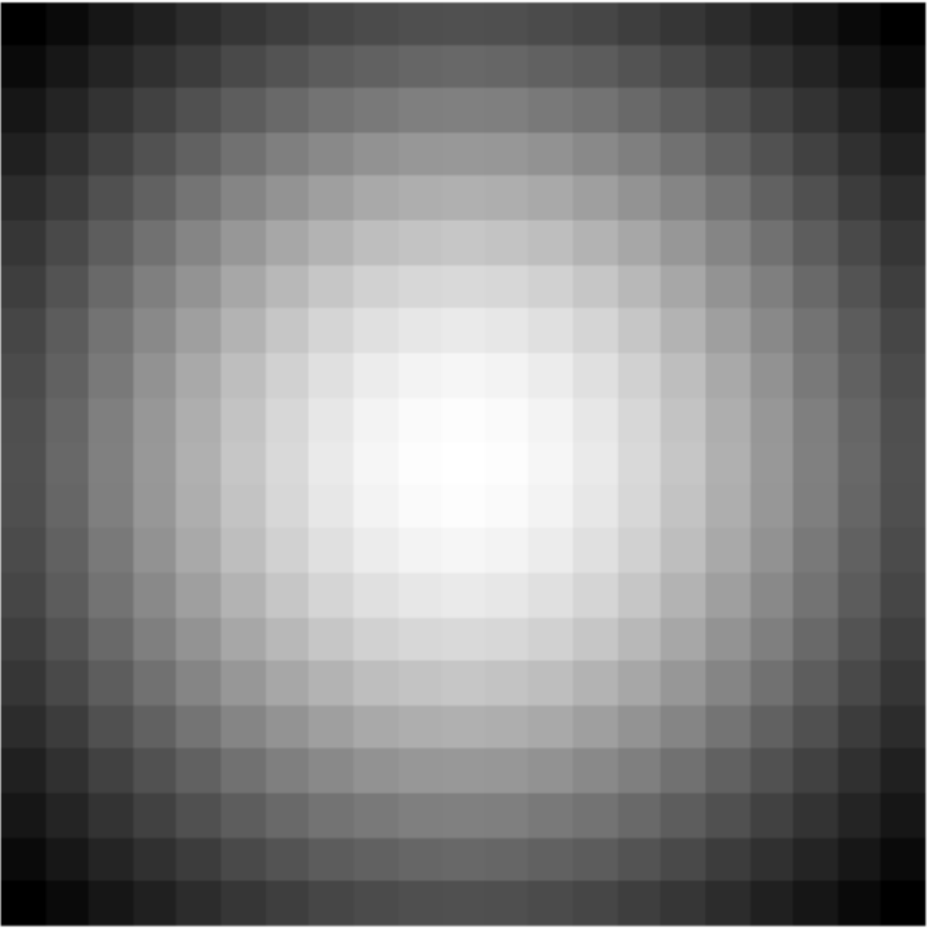}}
          \end{minipage}
          \tabularnewline
          \hline
          \hline
          \multirow{3}{*}{\tabincell{l}{RCAN \cite{zhang2018image}}}
          & 5 & 22.75 & 22.59 & 22.24 & 22.38 & 22.00 & 21.75 & 21.63 & 21.53 & 21.45
          \tabularnewline
          & 10 & 22.16 & 22.02 & 21.73 & 21.84 & 21.50 & 21.28 & 21.18 & 21.08 & 21.02
          \tabularnewline
          & 15 & 21.32 & 21.20 & 20.96 & 21.06 & 20.77 & 20.58 & 20.49 & 20.41 & 20.35       
          \tabularnewline
          \hline
          \multirow{3}{*}{\tabincell{l}{IKC \cite{gu2019blind}}}
          & 5 & 22.98 & 22.84 & 22.57 & 22.80 & 22.36 & 22.03 & 21.78 & 21.66 & 21.55
          \tabularnewline
          & 10 & 22.55 & 22.42 & 22.10 & 22.46 & 21.98 & 21.51 & 21.39 &  21.21 & 21.17
          \tabularnewline
          & 15 & 22.12 & 22.06 & 21.82 & 21.97 & 21.62 & 21.16 & 20.99 & 20.81 & 20.69
          \tabularnewline
          \hline
          \multirow{3}{*}{\tabincell{l}{DASR \cite{wang2021unsupervised}}}
          & 5 & {\color{gray}23.20} & {\color{gray}23.13} & {\color{gray}22.88} & {\color{gray}23.22} & {\color{gray}22.75} & {\color{gray}22.18} & {\color{gray}22.01} & {\color{gray}21.86} & {\color{gray}21.77}
          \tabularnewline
          & 10 & {\color{gray}23.18} & {\color{gray}23.05} & {\color{gray}22.76} & {\color{gray}23.09} & {\color{gray}22.58} & {\color{gray}22.09} & {\color{gray}21.93} & {\color{gray}21.80} & {\color{gray}21.71}
          \tabularnewline
          & 15 & {\color{gray}23.07} & {\color{gray}22.92} & {\color{gray}22.62} & {\color{gray}22.91} & {\color{gray}22.41} & {\color{gray}21.99} & {\color{gray}21.84} & {\color{gray}21.72} & {\color{gray}21.64}
          \tabularnewline
          \hline
          \multirow{3}{*}{CRL-SR (Ours)}
          & 5 & \textbf{23.27} & \textbf{23.19} & \textbf{22.95} & \textbf{23.36} & \textbf{22.93} & \textbf{22.39} & \textbf{22.17} & \textbf{22.01} & \textbf{21.90}
          \tabularnewline
          & 10 & \textbf{23.26} & \textbf{23.12} & \textbf{22.86} & \textbf{23.23} & \textbf{22.76} & \textbf{22.31} & \textbf{22.13} & \textbf{21.98} & \textbf{21.88}
          \tabularnewline
          & 15 & \textbf{23.13} & \textbf{22.99} & \textbf{22.73}& \textbf{23.06} & \textbf{22.60} & \textbf{22.22} & \textbf{22.05} & \textbf{21.91} & \textbf{21.81}
          \tabularnewline
          \hline
      \end{tabular}}
    \end{center}
    \vspace{-0.3pt}
  \end{table*}

\section{Conclusion}
%In this paper, we propose the CRL-SR for blind SR to utilize contrastive learning to handle diverse multi-modal and spatially variant degradations that are more general in the real-world images. Specifically, instead of performing degradation estimations, we introduce a novel design for contrastive decoupling encoding (CDE) to extract resolution-invariant features across HR and LR images and directly remove the degradation information of LR images. Furthermore, we propose a novel conditional contrastive loss to guide the contrastive feature refinement (CFR) to generate high-frequency details that are lost in image degradation. The evaluation on the synthetic datasets and real-world images demonstrates the effectiveness of our conditional contrastive loss and the two novel designs. Quantitative and qualitative experimental results show that our network achieves state-of-the-art performance for blind SR. In future work, we will further explore the potential of contrastive learning in feature generation and blind super-resolution with real degradations.
This paper presents CRL-SR, a blind SR technique that introduces contrastive learning to tackle multi-modal and spatially variant degradations in the real-world data. Instead of estimating degradation models and parameters, we design contrastive decoupling encoding which extracts resolution-invariant features across HR and LR images and removes the degradation information of LR images directly. In addition, we design contrastive feature refinement that introduces a conditional contrastive loss to guide to generate the lost high-frequency details. Experiments on synthetic and real-world images show that the proposed CRL-SR achieves superior SR performance quantitatively and qualitatively. Moving forward, we will continue to study contrastive learning in feature generation and blind SR with real-world degradation.

%\newpage
\small

\bibliography{CRL_SR}

\end{document}